\documentclass[letterpaper]{article} % DO NOT CHANGE THIS
\usepackage{aaai2026}  % DO NOT CHANGE THIS
\usepackage{times}  % DO NOT CHANGE THIS
\usepackage{helvet}  % DO NOT CHANGE THIS
\usepackage{courier}  % DO NOT CHANGE THIS
\usepackage[hyphens]{url}  % DO NOT CHANGE THIS
\usepackage{graphicx} % DO NOT CHANGE THIS
\usepackage{natbib}  % DO NOT CHANGE THIS AND DO NOT ADD ANY OPTIONS TO IT
\usepackage{caption} % DO NOT CHANGE THIS AND DO NOT ADD ANY OPTIONS TO IT
\usepackage{amsmath}
\usepackage{amsfonts}
\usepackage{multirow}
\usepackage{makecell}

\UseRawInputEncoding

\usepackage{algorithm}
\usepackage{algorithmic}
\usepackage{booktabs} 

\usepackage{newfloat}
\usepackage{listings}
\DeclareCaptionStyle{ruled}{labelfont=normalfont,labelsep=colon,strut=off} % DO NOT CHANGE THIS
\floatstyle{ruled}
\newfloat{listing}{tb}{lst}{}
\floatname{listing}{Listing}
\title{Yes FLoReNce, I Will Do Better Next Time! Agentic Feedback Reasoning for Humorous Meme Detection}
\author {
    Olivia Shanhong Liu\textsuperscript{\rm 1},
    Pai Chet Ng\textsuperscript{\rm 2},
    De Wen Soh\textsuperscript{\rm 1},
    Konstantinos N. Plataniotis\textsuperscript{\rm 3}
}
\affiliations {
    \textsuperscript{\rm 1}Information Systems Technology and Design Pillar, Singapore University of Technology and Design, Singapore\\
    \textsuperscript{\rm 2}Infocomm Technology Cluster, Singapore Institute of Technology, Singapore\\
    \textsuperscript{\rm 3}The Edward S. Rogers Sr. Department of Electrical and Computer Engineering, University of Toronto, Canada\\
    shanhong\_liu@mymail.sutd.edu.sg, paichet.ng@singaporetech.edu.sg, dewen\_soh@sutd.edu.sg,
    kostas@ece.utoronto.ca
}

\usepackage{bibentry}
\begin{document}

\maketitle

\begin{abstract}
% Humorous memes blend visual and textual cues to convey irony, satire, or social commentary, posing unique challenges for AI systems that must interpret intent rather than surface correlations. Existing multimodal or prompting-based models generate explanations for humor but operate in an open loop, lacking the ability to critique or refine their reasoning once a prediction is made.  
% We propose FLoReNce, an agentic feedback reasoning framework that treats meme understanding as a closed-loop process during learning and an open-loop process during inference. In the closed loop, a reasoning agent is critiqued by a judge; the error and semantic feedback are converted into control signals and stored in a feedback-informed, non-parametric knowledge base. At inference, the model retrieves similar judged experiences from this KB and uses them to modulate its prompt, enabling better, self-aligned reasoning without finetuning. On the PrideMM dataset, FLoReNce improves both predictive performance and explanation quality over static multimodal baselines, showing that feedback-regulated prompting is a viable path to adaptive meme humor understanding.

Humorous memes blend visual and textual cues to convey irony, satire, or social commentary, posing unique challenges for AI systems that must interpret intent rather than surface correlations. Existing multimodal or prompting-based models generate explanations for humor but operate in an open loop,
lacking the ability to critique or refine their reasoning once a prediction is made. We propose FLoReNce, an agentic feedback reasoning framework that treats meme understanding as a closed-loop process during learning and an open-loop process during inference. In the closed loop, a reasoning agent is critiqued by a judge; the error and semantic feedback are converted into control signals and stored in a feedback-informed, non-parametric knowledge base. At inference, the model retrieves similar judged experiences from this KB and uses
them to modulate its prompt, enabling better, self-aligned reasoning without finetuning. On the PrideMM dataset, FLoReNce improves both predictive performance and explanation quality over static multimodal baselines, showing that feedback-regulated prompting is a viable path to adaptive
meme humor understanding.

{\textcolor{red}{\textbf{Caution:} This paper contains offensive content due to the nature of the topic, which may be disturbing or offensive to some readers. Reader discretion is advised.}} 

% \textit{There is a thin line that separates laughter and pain, comedy and tragedy, humour and hurt.}\hfill―Erma Bombeck

% \begin{quote}
% ―Erma Bombeck
% \end{quote}
\end{abstract}


\section{Introduction}
\label{sec:intro}
Humorous memes play a central role in online discourse, shaping opinions and spreading social commentary through visual–textual wit \cite{li2024attention, xi2025multimodal}. Understanding their humor is not only crucial for applications such as content moderation, sentiment analysis, and cultural trend monitoring \cite{shifman2013memes, milner2018world, vasquez2021cats, rehman2025multimodal}, but also for building AI systems capable of interpreting human intent and nuance \cite{buaroiu2022automatic, kalloniatis2024computational}. However, humor in memes rarely resides in explicit features, it often emerges from subtle semantic interactions between image and text, such as irony, contrast, or metaphor. Existing classifiers that solely rely on correlations between pixels and words therefore fail to capture the deeper incongruity that defines humor \cite{rahman2025camfusion, singh2024ramm}. 

% To correctly judge whether a meme is humorous, a model must reason about meaning: relate textual statements to visual context, infer implicit contrasts, and reconcile them within cultural or social frames of reference. In short, humor detection is not a pattern-recognition problem but a \emph{reasoning} problem, requiring models to interpret, evaluate, and explain the underlying logic of the joke rather than merely label it \cite{attardo1994linguistic,raskin1985semantic,morreall1983taking,radford2021learning,alayrac2022flamingo,baichuan2023qwenvl}.
\begin{figure}[t]
  \centering
  % Put the exported image in ./figures (recommended)
  % Use PDF if possible for crisp text; PNG works too.
  \includegraphics[width=\columnwidth]{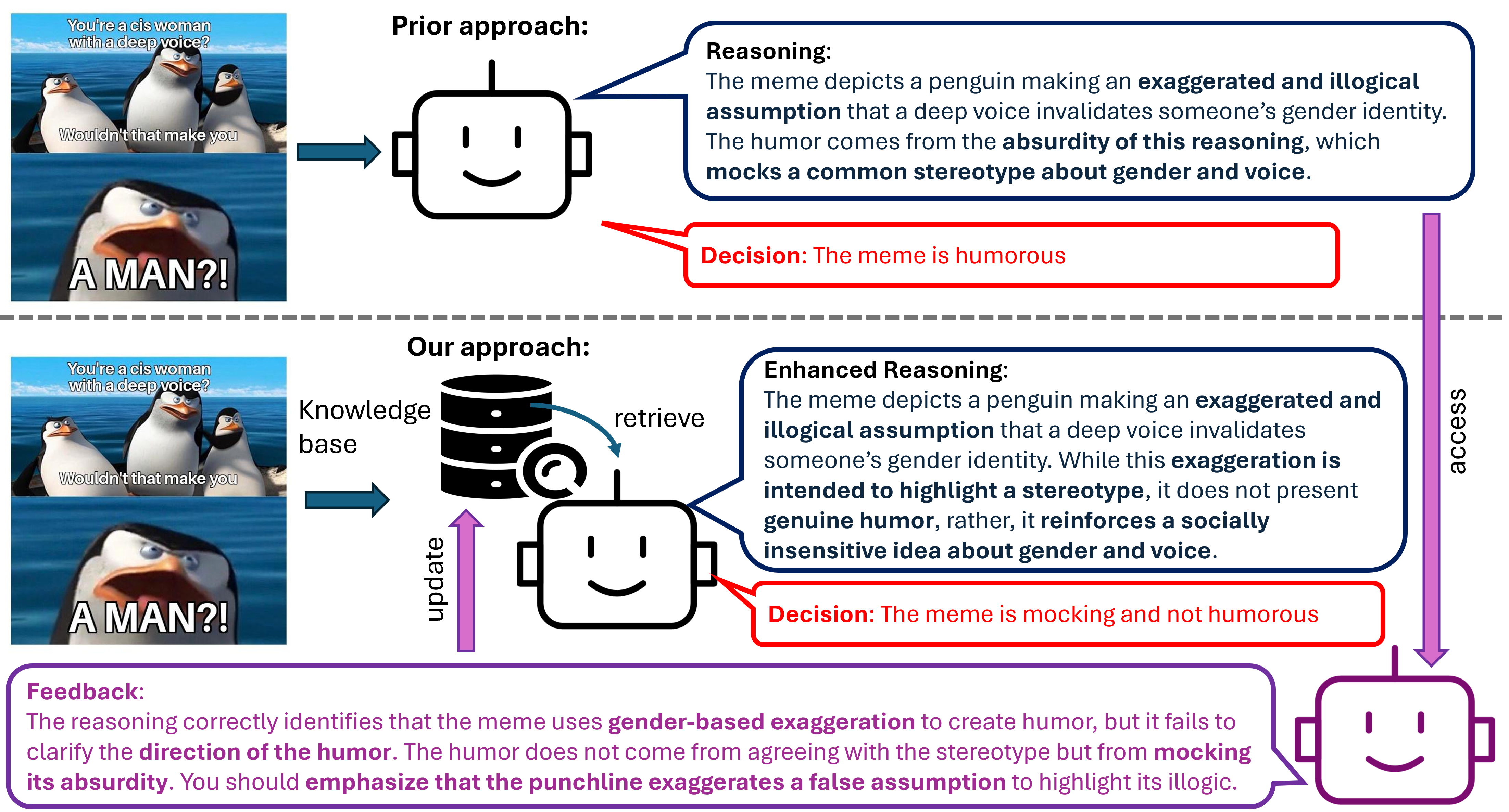}
  \caption{Comparison of our feedback-based reasoning correction approach (bottom) with prior approach (upper).
  }
  % The upper panel shows the initial reasoning process, interpreting the meme’s exaggerated logic as humorous due to its absurdity.
  % The lower panel demonstrates how the Judge Agent provides semantic feedback highlighting the misinterpretation of humor direction.
  % Through the feedback loop, the reasoning is refined
  \label{fig:feedback-correction}
  \vspace{-0.35cm}
\end{figure}

Recent studies have begun to incorporate reasoning into multimodal humor detection by prompting large vision–language models (VLMs) to generate textual explanations or multi-step justifications for their predictions. Approaches such as chain-of-thought prompting and multi-agent debate frameworks \cite{liu2023agentbench, madaan2023self, zong2024triad, li2024generation, schneider2025generative, liu2025breaking, zhang2025if, lee2025reliable} encourage models to articulate why an image–text pair may be humorous, improving interpretability over black-box classifiers like CLIP \cite{radford2021learning} or VL-BERT \cite{li2019visualbert}. Other methods, including MemeCLIP \cite{shah2024memeclip} and TCRNet \cite{kasu2025d}, attempt to fuse textual and visual cues for irony or incongruity reasoning. While these methods mark progress toward explainable humor understanding, they remain fundamentally static: once a model produces an incorrect or shallow rationale, there is no mechanism for correction, reflection, or adaptation based on feedback, as shown in the upper panel in Fig.~\ref{fig:feedback-correction}. In contrast, human humor comprehension is inherently dynamic, in which people refine their interpretations over time through critique, social feedback, and exposure to new cultural contexts, as illustrated in the lower panel of Fig.~\ref{fig:feedback-correction}. This feedback process stabilizes understanding much like a control system regulates output to minimize error. Without such a loop, AI reasoning tends to oscillate between over- and under-predicting humor, lacking the self-corrective mechanism that enables humans to adaptively adjust their interpretations.

In this paper, we propose \textbf{FLoReNce}, \textbf{F}eedback-\textbf{Lo}op \textbf{Re}asoner with \textbf{N}on-parametric Experien\textbf{ce} (FLoReNce), a novel agentic framework that addresses these limitations by introducing feedback-driven adaptation into multimodal humor understanding. Our FLoReNce models the reasoning process as a closed-loop system that continuously refines its interpretation through structured critique and control. It integrates four key modules: 1) a Vision–Language Reasoning Agent that interprets the meme and produces an initial rationale; 2) a Judge Agent that evaluates this reasoning against ground truth and issues semantic feedback; 3) a PID controller that transforms the prediction error and feedback into quantitative control signals; and 4) a Knowledge Base that accumulates these experiences as non-parametric memory. Together, these components enable FLoReNce to dynamically adjust its reasoning strategy at inference time, by retrieving similar past cases from memory and modulating its prompts based on prior feedback, without requiring any parameter updates or retraining. In doing so, FLoReNce bridges the gap between static reasoning and human-like adaptive understanding, allowing large vision–language models to learn from their own interpretive history and progressively stabilize their perception of humor.

Our key contributions are as follows:
\begin{itemize}
    \item We formulate humor reasoning as a closed-loop state-space system, where prediction errors and semantic feedback are treated as control signals regulating the reasoning process. This formulation enables control-driven adaptive prompting that dynamically adjusts the reasoning behavior of a frozen vision–language model for humor classification.
    \item We construct a feedback-informed Knowledge Base (KB) that encodes both the model’s own reasoning and the Judge’s critique, rather than storing raw training examples as in prior retrieval-based approaches. This non-parametric memory evolves through feedback, capturing interpretive refinements that improve inference-time reasoning.  
    \item We demonstrate the effectiveness of FLoReNce on the PrideMM dataset, achieving strong performance with enhanced interpretability and adaptability. Even with a minimal retrieval setting (\( \text{top-}K=1 \)), FLoReNce attains an F1-score of {0.7708}, showing that dynamic feedback and control substantially improve humor understanding over static reasoning baselines.  
\end{itemize}

\section{Related Works}

\subsection{Humorous Meme Detection}
Early research on humorous meme detection treated the task as a shallow multimodal classification problem.  
\citet{vlad2020upb} and \citet{guo2020guoym} explored fusing image and OCR text using handcrafted or CNN-based features.  
Later, transformer-based encoders such as BERT and ALBERT \cite{devlin2019bert, lan2019albert} were paired with deep visual backbones like VGG and DenseNet \cite{simonyan2014very, huang2017densely}, forming parallel branches or late-fusion pipelines \cite{gupta2020bennettnlp}.  
Subsequent work introduced cross-modal attention to align visual and textual cues for contextual humor understanding—e.g., \citet{pramanick2021exercise} used joint attention to capture image–text incongruity, and \citet{phan2022little, kumari2024mu2sts, singh2024well, kasu2025d} extended this with transformer-based multimodal fusion and multitask setups.  

Despite these advances, most existing models treat humor detection as a surface-level classification task, lacking the ability to reason about how or why a meme is humorous (Kumari et al. 2024; Liu et al. 2024). These models often conflate different humor mechanisms (e.g., irony, absurdity, wordplay) without modeling them explicitly. 

\begin{figure*}[th!]
    \centering
    \includegraphics[width=.95\linewidth]{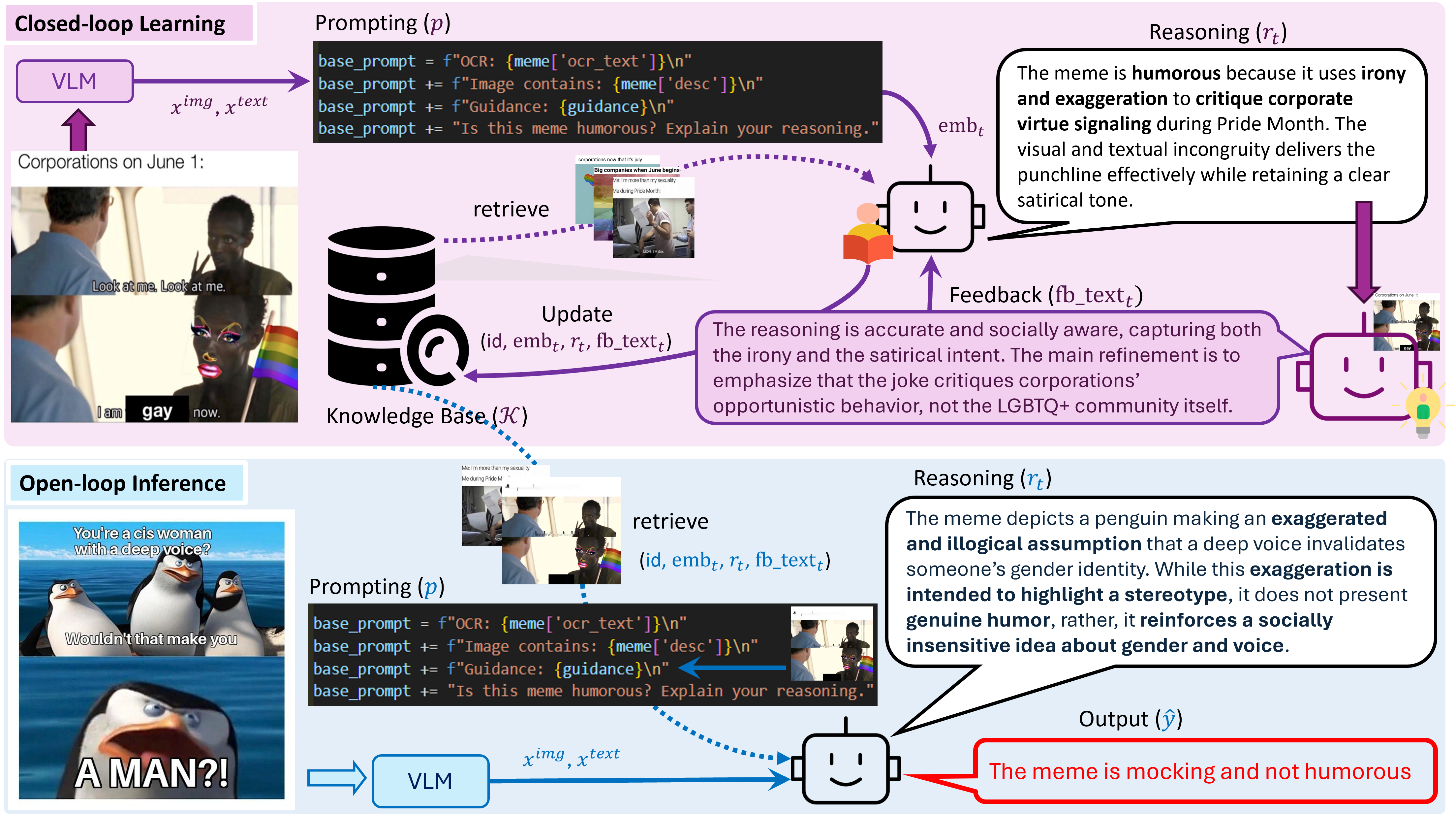}
    \caption{\textbf{FLoReNce framework.} 
    In the \emph{closed-loop learning} (top), the VLM reasons over memes, receives judge feedback, and stores judged embeddings in the Knowledge Base ($\mathcal{K}$). 
In the \emph{open-loop inference} (bottom), $\mathcal{K}$ provides retrieval-based guidance for adaptive prompting on unseen memes, distinguishing humorous from non-humorous cases.}
\label{fig:florence}
% \vspace{-0.3cm}
\end{figure*}
\subsection{LLM-Based Multi-Agent Frameworks}
Large Language Models (LLMs) have been increasingly adopted as autonomous agents capable of planning and deliberation across diverse domains such as embodied decision-making \cite{smit2023should, wang2023voyager, pham2023let}, collaborative problem solving \cite{liu2023agentbench, madaan2023self, zong2024triad, li2024generation}, and social simulation \cite{schneider2025generative, zhang2025if}.  
These efforts demonstrate that LLMs can reason beyond static prompting when embedded in multi-agent environments.  
Frameworks such as ChatEval \cite{chan2023chateval}, Debate \cite{liang2024encouraging, nguyen2025debate}, and CoT-SelfConsistency \cite{wang2022self} simulates interactive and consensus-driven reasoning, while stance and meme-understanding frameworks like COLA \cite{lan2024stance}, LoReHM \cite{huang2024towards}, and MiND \cite{liu2025mind} extend such agentic reasoning to multimodal social media analysis.  

However, these systems remain fundamentally open-loop: they exchange or aggregate textual responses but lack a principled feedback mechanism that regulates reasoning trajectories.  
Their refinement typically depends on oracle supervision or external evaluators, conditions infeasible in zero-shot or subjective classification settings such as humor or hate detection.  
{FLoReNce} departs from these paradigms by introducing a closed-loop control formulation in which judge feedback is numerically encoded, stored in a Knowledge Base, and reused to modulate future reasoning prompts.  
This bridges discrete linguistic critique with continuous control dynamics, transforming reactive multi-agent dialogue into a stable feedback-regulated reasoning process.
Prior works in multimodal humor understanding and LLM-based multi-agent reasoning either rely on static fusion or unregulated dialogue.  
FLoReNce unifies both strands under a control-theoretic perspective, where reasoning, judgment, and retrieval interact through feedback loops, offering a principled path toward interpretable and self-regulating humor understanding.

%%%%%%%%%%%%%%%%%%%%%%%%%%%%%%%%%%%%%%%%%%%%%%%%%%%%%%%%%%%%%%%%%%%%%%%%%%%%%%%%%%
\section{FLoReNce Framework}\label{sec:methodology}
% We propose  \textit{\textbf{F}eedback-\textbf{Lo}op \textbf{Re}asoner with \textbf{N}on-parametric Experien\textbf{ce}}, dubbed as FLoReNce, a closed-loop reasoning framework to simulate a corrective learning loop that enables the agent to improve its reasoning and prediction without model finetuning. As shown in Fig.~\ref{fig:florence}, FLoReNce replaces standard one-pass reasoning with a feedback-driven process involving two regimes:
% (i) \emph{Closed-loop learning}, where a Judge critiques the Reasoning Agent’s output; the error and critique are mapped by a controller into numeric control signals, and the system \emph{updates a feedback-informed knowledge base (KB)} that stores embeddings, rationales, and critiques; and
% (ii) \emph{Open-loop inference}, where the model \emph{retrieves} similar, judge-corrected experiences from the KB to form an adaptive control vector that modulates prompting and stabilizes reasoning at test time, again \emph{without} parameter updates.

\paragraph {Problem Statement.} We define a humorous meme detection dataset as a set of memes where each meme is denoted \(m=(x^{\mathrm{img}},x^{\mathrm{text}},y)\), with image \(x^{\mathrm{img}}\), OCR text \(x^{\mathrm{text}}\), and humor label \(y\in\{0,1\}\).
The \emph{Reasoning Agent} is a frozen VLM \(\mathcal{R}_\theta\), which, given a guidance prompt \(p\), outputs a humor score and a textual rationale; it also exposes a hidden representation used for retrieval.
We write this as
\begin{equation}
\begin{aligned}
    (\hat{y},\,r)=\mathcal{R}_\theta\!\big(x^{\mathrm{img}},x^{\mathrm{text}},p\big), \\
    \mathrm{emb}=\Phi_\theta\!\big(x^{\mathrm{img}},x^{\mathrm{text}},p\big)\in \mathbb{R}^d,
\end{aligned}
\end{equation}
where \(\hat y\in[0,1]\) is the predicted humor probability, \(r\) is the generated reasoning text, and \(\mathrm{emb}\) is the mean-pooled hidden embedding.

A numeric \emph{control vector} \(c\) is mapped to guidance text via a prompt-mapper
\begin{equation}
p=\Psi(c),
\end{equation}
which turns the control signal into succinct instructions (e.g., ``be conservative,'' ``check sarcasm,'' ``verify setup\(\rightarrow\)twist'') used by \(\mathcal{R}_\theta\).
Intuitively, \(\Psi\) is the interface that converts \emph{continuous} feedback into \emph{discrete, interpretable} prompting that steers the VLM's reasoning.

\subsection{Closed-Loop Learning}
\label{sec:closed-loop}
\paragraph{Agent reasoning.}
Given \(m\) and a current prompt \(p_t=\Psi(c_t)\), the agent produces \((\hat y_t,r_t)\) and an embedding \(\mathrm{emb}_t\) as above.
This step collects the agent's \emph{current belief} (score) and \emph{argument} (rationale) under the present guidance.

\paragraph{Judge feedback and control.}
The Judge \(\mathcal{J}_\phi\), with access to ground truth, critiques the agent by outputting a scalar error $e_t$, a textual critique $\text{fb\_text}_t$, and a low-dimensional feedback vector $f_t$:
\begin{equation}
\begin{aligned}
(e_t,\,\text{fb\_text}_t,\,f_t)=\mathcal{J}_\phi\!\big(m,\hat y_t,r_t\big),
\\
e_t = y - \hat y_t, \quad f_t\in\mathbb{R}^3.    
\end{aligned}
\end{equation}
Here \(e_t\) measures \emph{how far} the agent is from the target label, while \(f_t\) summarizes the critique semantics along three interpretable axes (e.g., irony/sarcasm, narrative structure, layout cues).
The Controller \(\mathcal{C}\) converts error history into a stabilizing action using PID dynamics:
\begin{equation}
\label{eq:pid}
u_t \;=\; K_P\,e_t \;+\; K_I \sum_{\tau=1}^{t} e_\tau \;+\; K_D \big(e_t - e_{t-1}\big).
\end{equation}
% Intuition: the \emph{proportional} term reacts immediately to current miscalibration; the \emph{integral} term cancels systematic bias accumulated over time (e.g., consistent over/under-detection of humor); the \emph{derivative} term damps rapid swings in judgment, improving stability of reasoning.
We aggregate numeric signals into a control vector
\begin{equation}
\label{eq:control-training}
c_t \;=\; \big[u_t,\ f_t^\top,\ k_t^\top\big]^\top \in \mathbb{R}^{1+3+3},
\end{equation}
where \(k_t\in\mathbb{R}^3\) is an optional compact KB signal (zero at the beginning of training).
The new prompt is then \(p_t=\Psi(c_t)\), which nudges the agent toward the aspects highlighted by the critique (e.g., ``check for sarcasm'' when \(f_{t,1}\) is large).

% %NOTE:
% scalar error $e_t$: drives numeric stabilization (PID) 
% textual critique $\text{fb\_text}_t$: is stored and human-interpretable 
% low-dimensional feedback vector $f_t$: provides a compact semantic control signal.

\paragraph{Experience-based KB update.}
After critique, we store \emph{feedback-informed} experience into a non-parametric KB:
\begin{equation}
\label{eq:kb-update}
\mathcal{K} \;\leftarrow\; \mathcal{K} \;\cup\; \big\{(id,\ \mathrm{emb}_t,\ r_t,\ \text{fb\_text}_t)\big\}.
\end{equation}
Unlike retrieval systems that index raw training pairs, \(\mathcal{K}\) preserves the \emph{reasoning trace} and the \emph{judge critique} alongside the embedding.
This makes later retrieval \emph{experience-aware}: the system recalls not only ``what it saw,'' but also ``how it was corrected.''

% \paragraph{State-space view.}
% Let \(x_t\) denote the (implicit) reasoning state of the agent, and \(y_t=h(x_t)\) the observable humor score.
% With judge feedback available in training, the closed-loop evolves as
% \begin{equation}
% \label{eq:ss-training}
% x_{t+1} \;=\; f\!\big(x_t,\,u_t,\,f_t,\,k_t\big),
% \qquad
% y_t \;=\; h(x_t),
% \qquad
% p_t \;=\; \Psi\!\big([u_t,\,f_t^\top,\,k_t^\top]^\top\big),
% \end{equation}
% where \eqref{eq:pid}--\eqref{eq:control-training} define how \(u_t\) and \(c_t\) are produced.
% Intuitively, \eqref{eq:ss-training} states that the agent's hidden reasoning state is updated by (i) a numeric stabilizer \(u_t\), (ii) semantic critique \(f_t\), and (iii) compact memory \(k_t\), while the observable \(y_t\) is the agent's current judgment.
% The prompt mapping \(\Psi\) forms the \emph{linguistic interface} that injects control into the VLM.

\subsection{Open-Loop Inference}
\label{sec:memory-driven}
\paragraph{Query embedding and retrieval.}
At test time there is \emph{no judge}: the system adapts by leveraging \(\mathcal{K}\) to shape prompts.
We first form a neutral/base prompt \(p_0=\Psi(\mathbf{0})\) and compute a query embedding
\begin{equation}
q \;=\; \Phi_\theta\!\big(x^{\mathrm{img}},x^{\mathrm{text}},p_0\big).
\end{equation}
We retrieve top-\(K\) neighbors by cosine similarity:
\begin{equation}
\label{eq:cosine}
\mathrm{sim}(q,\mathrm{emb}_j) \;=\; 
\frac{q^\top \mathrm{emb}_j}{\|q\|\,\|\mathrm{emb}_j\|},
\qquad
j \in \mathcal{N}_K(q).
\end{equation}
This step \emph{grounds} the current meme in past, judge-corrected experiences that were semantically similar.

\paragraph{Compact KB signal and control.}
We summarize retrieved entries into a compact memory signal
\begin{equation}
\label{eq:k-summarize}
k \;=\; \frac{1}{K}\sum_{j\in\mathcal{N}_K(q)} g\!\big(\mathrm{emb}_j\big) \;\in\; \mathbb{R}^3,
\end{equation}
where \(g:\mathbb{R}^d\!\to\!\mathbb{R}^3\) is a fixed projection used by the controller. %(code: \texttt{kb\_mix\_dim}=3)
With no judge present, we set \(f=\mathbf{0}\) and assemble
\begin{equation}
\label{eq:control-infer}
c \;=\; \big[u,\ f^\top,\ k^\top\big]^\top \;=\; \big[u,\ \mathbf{0}^\top,\ k^\top\big]^\top,
\end{equation}
where \(u\) is  a policy-driven scalar derived from \(k\) (e.g., more conservative when \(k\) signals risk of false positives).
The prompt is \(p=\Psi(c)\), which \emph{adapts} the agent's attention toward failure modes seen in similar past cases (e.g., stereotype inversion, sarcasm cues).

\paragraph{Final reasoning and inference.}
The agent produces the final judgment and rationale
\begin{equation}
(\hat y,\,r)=\mathcal{R}_\theta\!\big(x^{\mathrm{img}},x^{\mathrm{text}},p\big),
\end{equation}
completing a memory-driven loop without supervision.
% \begin{equation}
% \label{eq:ss-infer}
% x_{t+1} \;=\; f\!\big(x_t,\,u_t,\,\underbrace{0}_{f_t},\,k_t\big),
% \qquad
% y_t \;=\; h(x_t),
% \qquad
% p_t \;=\; \Psi\!\big([u_t,\,0^\top,\,k_t^\top]^\top\big).
% \end{equation}
% Equation~\eqref{eq:ss-infer} mirrors \eqref{eq:ss-training} but removes judge feedback, replacing it with experience retrieved from \(\mathcal{K}\).
Intuitively, the KB plays the role of a non-parametric prior over reasoning behavior: it nudges the agent toward historically successful interpretations for similar memes.
Because \(\mathcal{R}_\theta\) is frozen, this achieves inference-time adaptation not by changing weights, but by modulating prompts with control signals learned from prior feedback.

% \paragraph{Why this design works.}
% The PID term stabilizes the response to systematic miscalibration; the feedback vector encodes \emph{what} to check (semantic axes); and the KB signal encodes \emph{when/where} such checks mattered historically.
% The prompt mapper \(\Psi\) ties them together into human-readable instructions, ensuring that control remains \emph{interpretable}, \emph{modular}, and \emph{compatible} with frozen VLMs.

%%%%%%%%%%%%%%%%%%%%%%%%%%%%%%%%%%%%%%%%%%%%%%%%%%%%%%%%%%%%%%%%%%%%%%%%%%%%%%%%%%%%%%%%%%%%%%%%%%%%%%%%%%%%%%%%%%%%%%%%%
\section{Experiments Setup}
\subsection{Evaluation Datasets} 
We evaluate on PrideMM~\cite{shah2024memeclip}, a multimodal dataset of 5{,}063 text-embedded images (memes, posters, infographics) related to the LGBTQ+ movement, collected from Facebook, Twitter/X, and Reddit during 2020–2024. OCR text is extracted (with standard cleaning) and paired with the image.

Unless otherwise specified, we follow the split protocol used in the PrideMM paper’s experiments, i.e., a predefined {85/5/10} train/validation/test split. In our closed-loop learning stage, we build a feedback-informed KB on the training split without updating model weights: the Reasoning Agent is frozen, and the Judge accesses labels to issue critiques. We report final results on the held-out test split. At inference time, the Judge is disabled; the system performs retrieval from the KB and adaptive prompting without labels.

\begin{table*}[t]
\centering
\small
\caption{Results on PrideMM. Predictive Performance reports \emph{Accuracy}, \emph{Macro-F1}, \emph{MCC}, and \emph{RQ}. }
\vspace{-0.2cm}
\label{tab:pridemm_perf_rq}
\begin{tabular}{l l ccc c}
\toprule
& & \multicolumn{3}{c}{\text{Predictive Performance}} 
& \multicolumn{1}{c}{\makecell{\text{Reasoning Performance}}} \\
\cmidrule(lr){3-5}\cmidrule(l){6-6}
\text{Model} & \text{Backbone} & \text{Accuracy} & \text{Macro-F1} & \text{MCC} & \text{RQ (\%)} \\
\midrule
Visual Only                        & ResNet50 + MLP         & 66.08 & 61.67 & {0.33} & - \\
Text Only                          & T5 + MLP               & 67.85 & 66.10 & {0.36} & - \\
\midrule
MemeCLIP \citep{shah2024memeclip}  & CLIP                   & \textbf{78.30} & 76.99 & {0.57} & - \\
MOMENTA \citep{pramanick2021momenta} & CLIP                 & 73.57 & 69.92 & {0.47} & - \\
\midrule
PromptHate \citep{cao2022prompting} & RoBERTa               & 73.77 & 73.46 & {0.49} & - \\
LoReHM \citep{huang2024towards}     & LLaVA-34B             & 70.09 & 64.07 & {0.39} & {64.8} \\
COLA \citep{lan2024stance}          & GPT-3.5-Turbo         & 53.25 & 59.34 & {0.07} & {58.5} \\
MiND \citep{liu2025mind}            & Qwen2.5-VL-32B        & 54.45 & 50.43 & {0.05} & {52.6} \\
\midrule
\text{FLoReNce (K=1)}             & \multirow{4}{*}{\text{Qwen2.5-VL-32B}}
                                    & {73.40} & 77.08 & {0.48} & {74.0} \\
\text{FLoReNce (K=3)}             & 
                                    & 73.73 & \textbf{77.36} & {0.48} & {74.3} \\
\text{FLoReNce (K=5)}             & 
                                    & \underline{73.80} & \underline{77.33} & {0.48} & {74.4} \\
\text{FLoReNce (K=10)}            & 
                                    & {73.60} & \underline{77.33} & {0.47} & {74.2} \\
\bottomrule
\end{tabular}
\end{table*}
\subsection{Baselines}
We benchmark FLoReNce against a comprehensive set of text-only, vision-only, multimodal, and agentic/prompted baselines that collectively span the evolution of humor understanding in multimodal memes.

\paragraph{Classical and Multimodal Fusion.}
Early multimodal approaches treat humor as a feature-level fusion problem.  
{ResNet50+MLP} (visual-only) and {T5+MLP} (text-only) provide unimodal lower bounds.  
{MOMENTA} \cite{pramanick2021momenta} combines BERT-based textual features and ResNet-based visual features through multimodal transformers to detect image–text incongruity.  
{MemeCLIP} \cite{shah2024memeclip} leverages CLIP embeddings to align vision and text in a shared contrastive space and then trains a classifier for humor or hate recognition.  
These models encode multimodal correlation but lack explicit reasoning or feedback; their predictions are static once trained.

\paragraph{Prompt-based and Agentic Reasoning.}
Recent work explores LLM/VLMs as reasoning agents.  
{PromptHate} \cite{cao2022prompting} reformulates meme classification as a textual entailment problem using prompt templates over RoBERTa.  
{LoReHM} \cite{huang2024towards} fine-tunes the LLaVA-34B VLM with low-rank adapters to reason about humor and harmfulness through visually grounded instructions.  
{COLA} \cite{lan2024stance} introduces collaborative multi-agent stance reasoning using GPT-3.5-Turbo, extending chain-of-thought and debate prompting to generate stance-aware textual judgments.  
{MiND} \cite{liu2025mind} employs Qwen2.5-VL-32B with iterative self-reflection for meme interpretation, representing a state-of-the-art agentic reasoning baseline on PrideMM.

These methods exploit in-context or few-shot prompting and LLM-as-agent reasoning, yet they remain open-loop: no explicit control feedback regulates their reasoning trajectory once a prompt is issued.

\subsection{Evaluation Metrics}
\paragraph{Predictive performance.}
We report three thresholded metrics:
Accuracy, Macro-F1 (unweighted mean of per-class F1), and
Matthews Correlation Coefficient (MCC), a robust, single-number summary under class imbalance.
Unless stated otherwise, we binarize $\hat y\in[0,1]$ at $0.5$ (optionally calibrated on \texttt{val}).

\paragraph{Reasoning quality.}
Let $y\in\{0,1\}$ be the ground truth and $\hat y\in\{0,1\}$ the predicted label.
Denote $N_1$ and $N_0$ the number of positive and negative examples, and
$\mathrm{TP}$, $\mathrm{TN}$ the counts of correct predictions in each class.
We define a basic correctness-aligned reasoning score
\[
\mathrm{RQ}
\;=\;
\frac{1}{2}\Bigg(
\frac{\mathrm{TP}}{N_1}
\;+\;
\frac{\mathrm{TN}}{N_0}
\Bigg)
% \;=\;
% \text{BalancedAccuracy}.
\]

\subsection{Implementation Details}
\textbf{Reasoning Agent} $\mathcal{R}_\theta$: Qwen2.5-VL-32B-Instruct (frozen).

\noindent
\textbf{Judge} $\mathcal{J}_\phi$: same backbone in supervision mode; produces $(e_t, \text{fb\_text}_t, f_t)$ with $e_t=y-\hat y_t$ and $f_t\in\mathbb{R}^3$ from MiniLM \texttt{all-MiniLM-L6-v2} (first 3 components).

\noindent
\textbf{Controller} $\mathcal{C}$: PID with $(K_P,K_I,K_D)$; states $(\sum e_t, e_{t-1})$ persist across samples during closed-loop learning.

\noindent
\textbf{Knowledge Base} $\mathcal{K}$: stores $(\text{id}, \text{emb}, r, \text{fb\_text})$, where \texttt{emb} is the mean of the last hidden layer from Qwen-VL. Retrieval uses cosine similarity on CPU tensors; the KB is JSONL (reproducible, low memory). The prompt mapper $\Psi$ converts control into guidance.

\noindent
\textbf{Closed-Loop Learning}
For each train meme $m_t$, the agent produces $(\hat y_t, r_t)$ and $\text{emb}_t$; the Judge returns $(e_t, f_t)$; the Controller computes $u_t$; and $\mathcal{K}$ is augmented with $(\text{id}_t,\text{emb}_t,r_t,\text{fb\_text}_t)$. PID states $(\sum e_t, e_{t-1})$ are preserved across steps. No test items are added to $\mathcal{K}$, and no model weights are updated.

\noindent
\textbf{Open-Loop Inference}
For each test meme, we compute a query embedding $q_t$, retrieve top-$K$ neighbors from $\mathcal{K}$, and summarize them into $k_t\in\mathbb{R}^3$.
With Judge disabled ($f_t=\mathbf{0}$), we form $c_t = [u_t,\ \mathbf{0}^\top,\ k_t^\top]$ and generate a controlled prompt $\Psi(c_t)$ for final $(\hat y_t,r_t)$.
Unless specified, we set $K=\texttt{TOP\_K}$ from the config.

\noindent
\textbf{Hyperparameters and Reproducibility}
Unless stated, we set $(K_P,K_I,K_D)=(1.0,0.5,0.1)$. % and \texttt{TOP\_K}=\textbf{1} \,/\,\textbf{3} (see Sec.~\ref{sec:ablations}).
We cap generation at 128 tokens (rationales).
All experiments run on NVIDIA L40S (48GB).
% We report the mean$\pm$std over three seeds \{13, 17, 23\}.
% KB size (train): \#entries = XXXX; retrieval latency (median): YY ms; peak RAM: ZZ GB.

% \begin{table*}[t]
% \centering
% \small
% \caption{Results on PrideMM (Accuracy and Macro-F1, in \%).}
% \label{tab:pridemm_memotion}
% \resizebox{\textwidth}{!}{%
% \begin{tabular}{l|l||cc}
% \toprule
% \multicolumn{2}{c}{\text{Dataset}}  & \multicolumn{2}{c}{\text{PrideMM}} \\
% \midrule
% \text{Model} &LM   & \text{Accuracy} & \text{Macro-F1} \\
% \midrule
%  Visual Only  &Resnet50 + MLP           & 66.08 & 61.67   \\
% Text Only   &T5 + MLP             & 67.85 & 66.10  \\
% \midrule
%  MemeCLIP \citep{shah2024memeclip}    &CLIP    & 78.30  & 76.99  \\
%  MOMENTA \citep{pramanick2021momenta}    &CLIP          & 73.57 & 69.92  \\
% \midrule
% PromptHate \citep{cao2022prompting} &RoBERTa  & 73.77 & 73.46 \\
% LoReHM \citep{huang2024towards}    &LLaVA-34B     & 70.09 & 64.07  \\
% COLA \citep{lan2024stance} &GPT-3.5-Turbo   & 53.25 & 59.34 \\
% MiND \citep{liu2025mind}   &Qwen2.5-VL-32B       & 71.75 & 70.23  \\
% \midrule
% FLoReNce (Ours)   &Qwen2.5-VL-32B      & 73.73 & 70.86 \\
% \bottomrule
% \end{tabular}%
% }
% \end{table*}

\begin{figure*}[th!]
    \centering
    \includegraphics[width=1\linewidth]{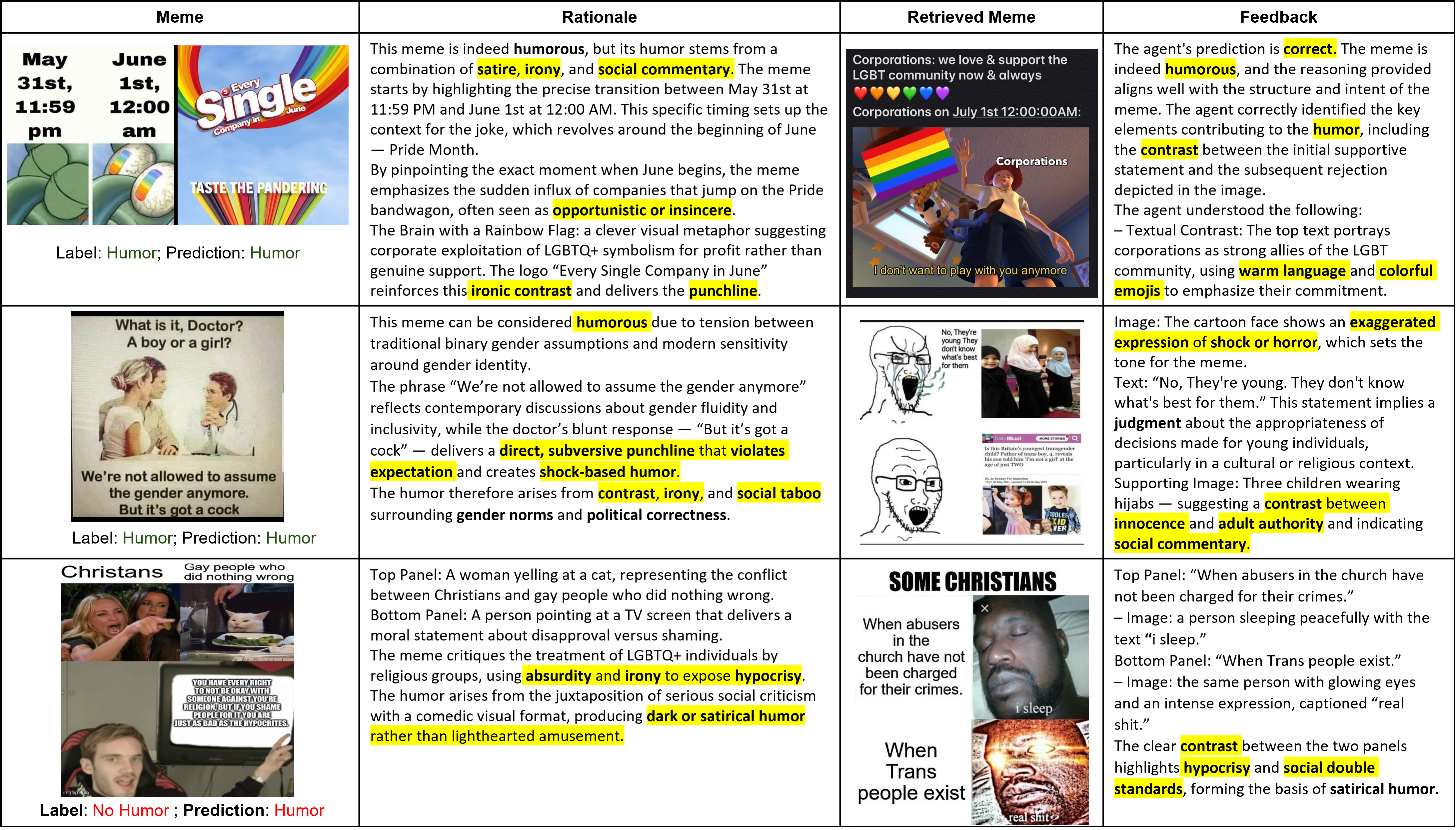}
    \caption{Examples of retrieved memes with feedbacks. Correct prediction in \textcolor{green}{green}. Incorrect prediction in \textcolor{red}{red}.}
    \label{fig:case}
\end{figure*}

\section{Results and Discussions}
The results in Table~\ref{tab:pridemm_perf_rq} show that FLoReNce attains predictive performance on PrideMM that is comparable to, and in several cases slightly better than, strong multimodal baselines. Classical unimodal systems (Visual Only, Text Only) remain in the mid-\(60\%\) range for accuracy, confirming that humor in LGBTQ\(+\) memes is genuinely multimodal. Established multimodal approaches such as MemeCLIP reach higher accuracy (78.30\%) and macro-F1 (76.99\%), indicating that better image-text alignment helps detect incongruity. Prompt-based or agentic baselines (PromptHate, LoReHM, MiND) mostly fall in the low-to-mid \(70\%\) accuracy range, with macro-F1 between \(64\%\) and \(73\%\). Our framework, FLoReNce with retrieval and control at \(K=3\), reaches 73.73\% accuracy and 77.36\% macro-F1, which is notable because the macro-F1 gain is larger than the gain in accuracy. This suggests that the feedback-informed KB and control signal are especially helpful for the harder class, improving class-balanced performance rather than only the dominant class.

Across the FLoReNce configurations (K=1, 3, 5, 10), the performance is remarkably stable: accuracy remains around 73--74\%, macro-F1 stays around 77\%, and the reasoning-quality score hovers near 74\%. Since RQ is defined in a balanced-accuracy style over correctly reasoned instances, this pattern implies that once the KB has been populated in the closed-loop phase, the open-loop inference can exploit even a small number of retrieved neighbours (K=1 or 3) to produce consistent, feedback-aligned reasoning. Increasing K beyond 3 does not deteriorate performance, which indicates that the retrieved experiences are semantically coherent and that the prompt-mapping function \(\Psi\) can absorb slightly richer control signals. Overall, these findings support the central claim of FLoReNce: integrating control-style feedback with a non-parametric, feedback-built memory improves not only raw prediction but also the stability and consistency of humor reasoning.

\subsection{Case Study}
Figure~\ref{fig:case} illustrates three representative examples from PrideMM showing how FLoReNce leverages feedback and retrieval for adaptive humor reasoning.  
In the first two cases (top rows), the system successfully classifies the memes as \emph{humorous}.  
For the “\textit{Corporations on June 1}” meme, the retrieved example from the KB contains prior judge feedback emphasizing irony, contrast, and corporate opportunism, which guides the Reasoning Agent to identify the satirical intent and deliver a socially aware explanation.  
Similarly, in the “\textit{Doctor / Gender}” meme, retrieval provides feedback cues such as subversive punchline, shock-based humor, and social taboo, helping the agent contextualize the blunt dialogue as intentional irony rather than literal bias.  
These examples demonstrate how feedback-informed retrieval enables FLoReNce to internalize semantic corrections, transforming judge supervision into reusable reasoning guidance.  
The third case (“\textit{Christians vs. Gay People}”) highlights a failure: although the meme was labeled non-humorous, the system predicted humorous because retrieved feedback on similar religious-context memes overemphasized absurdity and irony, causing misalignment between form (satirical tone) and intent (mocking content).  
This case underlines the challenge of distinguishing satire that critiques power from mockery that targets marginalized groups, a boundary that FLoReNce aims to learn more robustly through future refinement of judge feedback and retrieval filtering.

\subsection{Ablation Studies}
To understand the contribution of each component in FLoReNce, we progressively removed the KB, the control path, and the judge-derived semantic feedback \(f_t\), while keeping the retrieval size fixed at \(K=3\).
The plain VLM only achieve mid-range performance on PrideMM, while adding the KB alone yields a considerable gain, showing that retrieving feedback-informed experiences helps even without extra control.
Adding only the controller also improves over the base model, though slightly less than KB-only, indicating that control is more effective when it can condition on meaningful memory signals.
When we keep PID and KB but drop the semantic feedback vector \(f_t\), performance decreases compared to the full model, which confirms that judge critiques carry information that cannot be recovered from embeddings alone.
The best results are obtained when all three ingredients are present (PID + KB + \(f_t\)), supporting our claim that humorous meme understanding benefits from a closed-loop design that fuses numeric control, semantic feedback, and non-parametric memory.

\begin{table}[t]
\centering
\small
\caption{Component ablation on PrideMM (K=3).}
\label{tab:ablate_components}
\resizebox{\columnwidth}{!}{%
\begin{tabular}{lccc c}
\toprule
\text{Variant} & \text{Acc} & \text{Macro-F1} & \text{MCC}  \\
\midrule
Base VLM (no KB, no control)            & 64.20 & 58.10 & 0.22  \\
+ KB only (no control)                  & 68.30 & 63.90 & 0.35  \\
+ Control only (no KB)                  & 72.00 & 69.40 & 0.44  \\
-- $f_t$ (PID+KB, drop feedback vec)    & 73.00 & 70.20 & 0.46  \\
-- PID (KB signal only)                 & 72.60 & 70.00 & 0.45  \\
\textbf{Full FLoReNce (PID+KB+$f_t$)}   & \textbf{73.73} & \textbf{77.36} & \textbf{0.48} \\
\bottomrule
\end{tabular}}
\end{table}

\section{Conclusion}
% Humour is subjective, and there is subjectiveness in human annotations. 
In this paper, we delved into multimodal humor understanding and proposed FLoReNce, which treats the task as a regulated, experience-aware process rather than a one-shot classification problem. We introduced a feedback-loop formulation that transforms judge critiques into control signals and stores them as non-parametric experience, enabling iterative refinement of both predictions and rationales. On PrideMM, this yields improvements in accuracy and in the stability and consistency of generated reasonin even under minimal retrieval, showing that feedback-informed prompting can be a practical alternative to full model fine-tuning for subjective, nuance-heavy phenomena. Collectively, these findings position FLoReNce as a general recipe for controllable, critique-driven reasoning in multimodal settings, and open avenues for richer memory design.

\bibliography{aaai2026}

\end{document}